\documentclass[review,3p,times,twocolumn]{elsarticle}

\journal{}

\usepackage{amsmath,amssymb,amsfonts}
\usepackage{algorithmic}
\usepackage{algorithm}
\usepackage{graphicx}
\usepackage{textcomp}
\usepackage{xcolor}
\usepackage{physics}
\usepackage{tikz}
\usepackage{siunitx}
\usepackage{array}
\usepackage{multirow}
\usepackage{tabularx}
\usepackage{booktabs}
\usepackage{float}
\usepackage{bm}
\usepackage{makecell}
\usepackage{url}
\usepackage{hyperref}

\newtheorem{proposition}{Proposition}

\newtheorem{remark}{Remark}
\newdefinition{defn}{Definition}

\begin{document}

\begin{frontmatter}

\title{Collaborative Temporal Feature Generation via Critic-Free Reinforcement Learning for Cross-User Sensor-Based Activity Recognition}

\author[nuist-ai]{Xiaozhou~Ye\corref{cor1}}
\ead{200102@nuist.edu.cn}
\author[nuist-ai]{Feng~Jiang}
\author[nuist-cs]{Zihan~Wang}
\author[nuist-nj,nuist-ai]{Xiulai~Wang\corref{cor2}}
\ead{900231@nuist.edu.cn}
\author[nuist-nj]{Yutao~Zhang\corref{cor2}}
\ead{003448@nuist.edu.cn}
\author[auckland]{Kevin~I-Kai~Wang}

\cortext[cor1]{First author.}
\cortext[cor2]{Corresponding author.}

\affiliation[nuist-ai]{organization={School of Artificial Intelligence (School of Future Technology), Nanjing University of Information Science and Technology},
            city={Nanjing},
            country={China}}

\affiliation[nuist-cs]{organization={School of Cyberspace Security, Nanjing University of Information Science and Technology},
            city={Nanjing},
            country={China}}

\affiliation[nuist-nj]{organization={Jinling Hospital, Affiliated Hospital of Medical School, Nanjing University},
            city={Nanjing},
            country={China}}

\affiliation[auckland]{organization={Department of Electrical, Computer, and Software Engineering, The University of Auckland},
            city={Auckland},
            country={New Zealand}}

\begin{abstract}
Human Activity Recognition using wearable inertial sensors is foundational to healthcare monitoring, fitness analytics, and context-aware computing, yet its deployment is hindered by cross-user variability arising from heterogeneous physiological traits, motor habits, and sensor placements. Existing domain generalization approaches---including adversarial alignment, causal invariance learning, and contrastive pretraining---either neglect temporal dependencies in sensor streams or depend on impractical target-domain annotations. We propose a fundamentally different paradigm: modeling generalizable feature extraction as a collaborative sequential generation process governed by reinforcement learning. Our framework, \textbf{CTFG} (Collaborative Temporal Feature Generation), employs a Transformer-based autoregressive generator that incrementally constructs feature token sequences, each conditioned on prior context and the encoded sensor input. The generator is optimized via Group-Relative Policy Optimization, a critic-free algorithm that evaluates each generated sequence against a cohort of alternatives sampled from the same input, deriving advantages through intra-group normalization rather than learned value estimation. This design eliminates the distribution-dependent bias inherent in critic-based methods and provides self-calibrating optimization signals that remain stable across heterogeneous user distributions. A tri-objective reward comprising class discrimination, cross-user invariance, and temporal fidelity jointly shapes the feature space to separate activities, align user distributions, and preserve fine-grained temporal content. Evaluations on the DSADS and PAMAP2 benchmarks demonstrate state-of-the-art cross-user accuracy (88.53\% and 75.22\%), substantial reduction in inter-task training variance, accelerated convergence, and robust generalization under varying action-space dimensionalities.
\end{abstract}


\end{frontmatter}

\section{Introduction}
\label{sec:introduction}

Wearable inertial measurement units (IMUs) capture the kinematic signatures of human movement as multivariate time series, enabling automatic classification of activities such as walking, cycling, and stair climbing~\cite{ye2025cross, haresamudram2025transfer}. These Human Activity Recognition (HAR) systems underpin clinical gait assessment, fall detection in elderly care, adaptive fitness coaching, and smart-environment interaction~\cite{islam2023multilevel}. Despite significant progress in deep learning for HAR, a persistent challenge remains: models trained on a cohort of source users exhibit substantial performance degradation when deployed on unseen target users. This cross-user distribution shift is rooted in physiological diversity---limb length governs stride kinematics, body mass modulates ground-reaction-force profiles, and muscle composition alters the spectral characteristics of accelerometer signals---causing even identical activities to produce markedly different inertial signatures across individuals~\cite{barshan2025bidirectional, qin2023affar}.

Domain generalization (DG) addresses this challenge by learning representations from labeled source domains that transfer to arbitrary unseen targets without any target data during training~\cite{zhou2022domain}. The field has seen rapid progress: causal invariance learning discovers features robust to environment shifts through concept-level disentanglement~\cite{xiong2025categorical, shao2025cargi}; contrastive pretraining aligns representations across datasets and device configurations via self-supervised objectives~\cite{hong2024crosshar, yoon2024contrastsense}; adversarial domain generalization synthesizes diverse pseudo-domain features to bridge distributional discrepancies~\cite{liu2025gadpn, wang2026msdgm}; and time-series out-of-distribution frameworks partition latent subdomains to capture distribution heterogeneity~\cite{lu2024diversify}. Despite these advances, three recurrent limitations persist in the context of cross-user HAR.

Adversarial alignment methods operate on aggregated or frame-level feature representations, discarding the sequential structure that encodes activity dynamics---the precise feature class that is most stable across users~\cite{suh2023tasked}. Causal deconfounding approaches~\cite{xiong2025deconfounding} disentangle domain-specific and domain-invariant factors but do not exploit the multi-scale temporal relational structure intrinsic to human motion. Human activities are compositionally structured across time, with discriminative information residing in the evolving relationships between temporal phases (e.g., the stance-to-swing transition in gait, the extension-flexion cycle in pedaling) rather than in individual time steps~\cite{wen2023transformers_timeseries}. Conventional supervised objectives perform single-pass feature extraction that conflates this compositional temporal structure into monolithic representations, making it difficult to disentangle user-invariant dynamics from user-specific artifacts~\cite{essa2023temporal}. Furthermore, meta-learning approaches require domain-specific labels and per-user calibration~\cite{wang2024optimization_free}, while data augmentation strategies generate synthetic diversity that often introduces artifacts rather than capturing temporal invariances~\cite{zhang2024diverse_kdd}. Source-free adaptation reduces deployment overhead but still assumes access to a pretrained source model~\cite{kang2024sfadapter}.

To address these limitations, we propose to reformulate feature extraction as an active, multi-step generation process governed by reinforcement learning (RL), where a policy network constructs feature representations incrementally---token by token---with each decision conditioned on prior context. Two structural observations motivate this formulation. If the discriminative information in activity signals is distributed across temporal phases, then the feature extraction process should mirror this structure: an autoregressive generator that builds tokens sequentially can first capture coarse temporal motifs (overall periodicity) and then progressively refine them with finer details (phase relationships, transition dynamics). Moreover, the desirable properties of a generalizable feature space---inter-class separation and intra-class cross-user alignment---are distributional properties of the complete representation, not of individual tokens. RL provides a natural mechanism for evaluating these properties: the reward is computed over the entire generated token sequence, enabling holistic assessment that per-token surrogate losses cannot provide~\cite{sutton1998reinforcement}.

While RL-based feature generation addresses the optimization-representation mismatch, standard policy gradient methods such as Proximal Policy Optimization (PPO)~\cite{schulman2017proximal} require a learned value function to estimate per-step advantages. In the cross-user HAR context, this value function must estimate expected returns across feature distributions generated from heterogeneous source users, yet it is itself trained on source data and susceptible to the distribution biases the framework seeks to overcome. Group-Relative Policy Optimization (GRPO)~\cite{guo2025deepseekr1_nature}, resolves this by eliminating the value function. For each input, GRPO samples a group of candidate feature sequences, computes rewards for each, and derives advantages by normalizing against the group's statistics. This collaborative evaluation---where each generation is assessed relative to contemporaneous alternatives---provides three structural benefits: (1) the advantage signal is self-calibrating and invariant to absolute reward scale~\cite{ahmadian2024rloo}; (2) the group sampling inherently encourages exploration of diverse feature configurations~\cite{yu2025dapo}; and (3) eliminating the critic network reduces the parameter surface susceptible to source-user overfitting.

In summary, we introduce \textbf{CTFG} (Collaborative Temporal Feature Generation), a framework integrating autoregressive feature generation, critic-free policy optimization, and a tri-objective reward system for cross-user HAR. The key contributions of this paper are:

\begin{enumerate}
\item[(1)] We reformulate domain-generalizable feature extraction as collaborative sequential generation, where a Transformer-based autoregressive policy constructs temporal token sequences optimized via GRPO---a critic-free RL algorithm that derives advantages through intra-group reward normalization rather than learned value estimation. We provide formal analysis (Proposition~\ref{prop:variance}) showing that group-relative advantages yield affine-invariant gradient signals that remain stable under the reward-scale variation induced by heterogeneous source-user distributions, resolving the value-function bottleneck inherent in critic-based RL for cross-user HAR.

\item[(2)] We design an autoregressive Transformer decoder that generates feature tokens incrementally, with each token conditioned on the encoded sensor input and all previously generated tokens via causal self-attention and cross-attention. This compositional construction captures the hierarchical temporal structure of human activities---from coarse periodicity (gait cycles, pedaling rhythm) to fine-grained phase transitions (stance-to-swing, extension-to-flexion)---producing representations where user-invariant temporal dynamics are explicitly separated from user-specific amplitude artifacts.

\item[(3)] We propose a tri-objective reward mechanism combining class discrimination, cross-user invariance, and temporal fidelity. The temporal fidelity component, novel in the context of RL-driven HAR, explicitly penalizes information loss during feature generation by requiring reconstructibility of the encoder's temporal representation, preventing the policy from discarding fine-grained temporal content in pursuit of distributional alignment objectives.
\end{enumerate}

The remainder of this paper is organized as follows. Section~\ref{sec:related} reviews related work. Section~\ref{sec:methodology} presents the methodology with formal analysis. Section~\ref{sec:experiments} reports experimental results. Section~\ref{sec:conclusion} concludes with future directions.

\section{Related work}
\label{sec:related}

\subsection{Generalizable representations for wearable sensor HAR}

Domain generalization in HAR aims to build models from multiple source users that generalize to unseen targets without target data~\cite{haresamudram2025transfer, zhou2022domain}. Recent advances span several paradigms.

\textbf{Causal and invariance-based methods.} Xiong et al.~\cite{xiong2025categorical} proposed categorical concept invariant learning, separating causal activity features from domain-specific confounders via concept-level disentanglement. The same group introduced a deconfounding causal inference framework using two-branch architectures with early forking to isolate domain-invariant features via the Hilbert-Schmidt Independence Criterion~\cite{xiong2025deconfounding}. Shao et al.~\cite{shao2025cargi} bridged single domain generalization to causal concepts, proposing CaRGI, a framework that employs generative intervention to enlarge domain shifts with semantic consistency while learning causal representations via counterfactual inference. Lu et al.~\cite{lu2024diversify} presented DIVERSIFY method, a general framework for time-series out-of-distribution detection and generalization that discovers latent subdomains through iterative adversarial partitioning. While these methods achieve principled distribution-shift robustness, they operate on aggregate representations and do not exploit the sequential temporal structure intrinsic to human motion.

\textbf{Contrastive and self-supervised approaches.} CrossHAR~\cite{hong2024crosshar} achieves cross-dataset generalization through hierarchical self-supervised pretraining with sensor-aware augmentation. ContrastSense~\cite{yoon2024contrastsense} employs domain-invariant contrastive learning with parameter-wise penalties for in-the-wild wearable sensing. UniMTS~\cite{zhang2024unimts} introduced unified pre-training for motion time series, aligning sensor data with LLM-enriched text descriptions. DDLearn~\cite{qin2023ddlearn} combines distribution divergence enlargement with supervised contrastive learning for low-resource generalization. These approaches produce powerful pretrained representations but require large-scale pretraining corpora and lack mechanisms for explicit temporal structure preservation during feature extraction.

\textbf{Domain adaptation approaches.} Adaptive Feature Fusion (AFFAR)~\cite{qin2023affar} dynamically fuses domain-invariant and domain-specific representations. SWL-Adapt~\cite{hu2023swladapt} uses meta-optimization-based sample weight learning for cross-user adaptation. DWLR~\cite{ijcai2024dwlr} addresses the compound challenge of feature shift and label shift across users. SF-Adapter~\cite{kang2024sfadapter} enables computational-efficient source-free adaptation for edge deployment. Liu et al.~\cite{liu2025gadpn} introduced GADPN, a generative adversarial domain-enhanced prediction network that synthesizes diverse pseudo-domain samples to mitigate distributional discrepancies in multi-source domain generalization scenarios. Wang et al.~\cite{wang2026msdgm} proposed MSDGM, a lightweight multi-source domain generalization model combining MobileNet-based feature extraction with Mamba-based dynamic parameter adjustment, demonstrating effective generalization across unseen domains. However, adaptation methods require unlabeled target data, violating the strict DG constraint where target users are entirely unknown during training.

\textbf{Transformer-based cross-subject methods.} TASKED~\cite{suh2023tasked} combines Transformer architecture with adversarial learning and self-knowledge distillation for cross-subject HAR. CapMatch~\cite{xiao2025capmatch} introduces semi-supervised contrastive Transformer capsules with knowledge distillation. These extensions advance cross-user generalization but inherit limitations from adversarial training (instability, mode collapse) and treat sensor data as fixed-window snapshots rather than sequential processes.

\subsection{Policy optimization: from critic-based to critic-free methods}

PPO~\cite{schulman2017proximal} is the dominant policy gradient algorithm, using advantages estimated via Generalized Advantage Estimation (GAE)~\cite{schulman2016high} from a learned value function. Recent work in language model alignment has demonstrated that critic-free alternatives provide comparable or superior performance. Ahmadian et al.~\cite{ahmadian2024rloo} introduced REINFORCE Leave-One-Out (RLOO), estimating value baselines from multiple sampled completions. ReMax~\cite{li2024remax} eliminates the critic using a max-reward baseline. Direct Preference Optimization (DPO)~\cite{rafailov2023dpo} reparameterizes the RL objective into a supervised loss, bypassing both reward model and critic.

GRPO~\cite{guo2025deepseekr1_nature}, computes advantages by sampling multiple outputs per input and normalizing their rewards within the group. DAPO~\cite{yu2025dapo} extends this paradigm with clip-higher mechanisms and dynamic sampling. Guan et al.~\cite{guan2024rfdg} demonstrated that RL-based sample reweighting can serve federated domain generalization, though without autoregressive generation. To our knowledge, no prior work has applied GRPO---or any critic-free RL method---to sequential feature generation for sensor-based domain generalization.

\subsection{Temporal modeling for sensor data}

Transformers have advanced time-series modeling by enabling global temporal reasoning via self-attention, overcoming vanishing-gradient limitations of RNNs and fixed receptive fields of CNNs~\cite{wen2023transformers_timeseries}. ConvTran~\cite{foumani2024convtran} improved positional encoding for multivariate time series classification. Essa and Abdelmaksoud~\cite{essa2023temporal} combined temporal-channel convolutions with self-attention for HAR, while Ek et al.~\cite{ek2023transformer_heterogeneous} demonstrated Transformer robustness to device heterogeneity. Islam et al.~\cite{islam2023multilevel} presented multi-level attention-based feature fusion for multimodal wearable HAR in Information Fusion. IMUGPT~2.0~\cite{leng2024imugpt2} leveraged LLMs for cross-modality IMU data generation. However, existing Transformer-based HAR models use supervised objectives that do not explicitly promote user-invariant representations, and autoregressive variants are designed for prediction rather than generalizable feature construction.

Our framework bridges this gap by situating the Transformer within an RL loop where autoregressive generation becomes the mechanism for exploring and constructing generalizable representations, guided by distributional reward signals that enforce cross-user consistency through a critic-free optimization paradigm.

\section{Methodology}
\label{sec:methodology}

\begin{figure*}[!t]
\centering
\includegraphics[width=\textwidth]{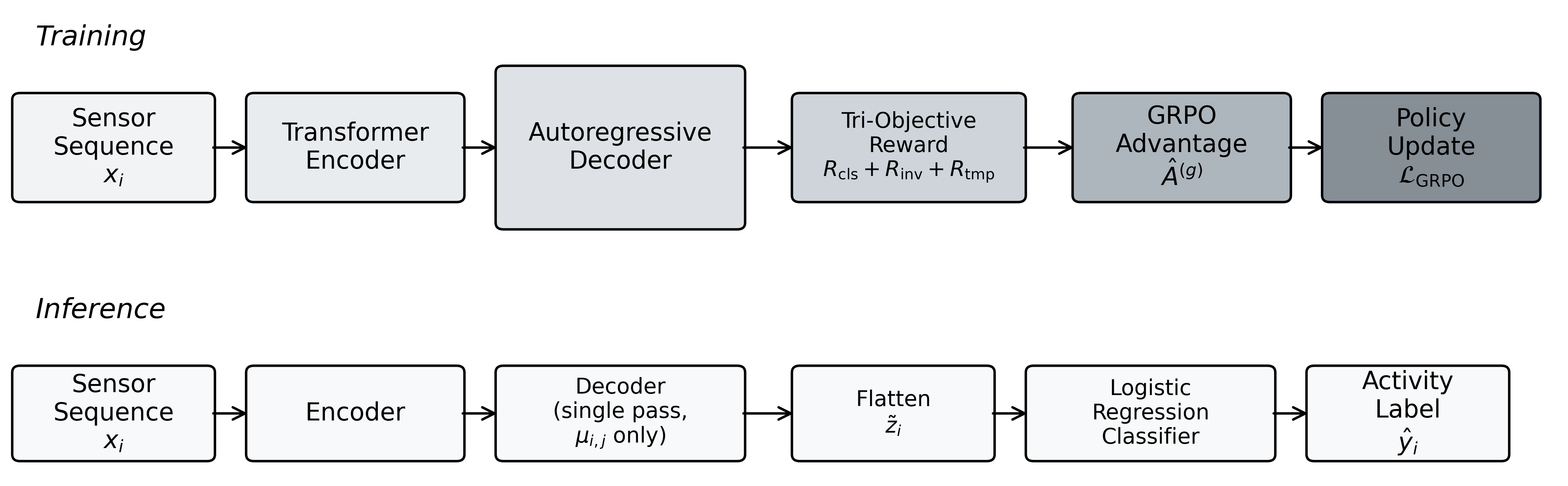}
\caption{Overview of the proposed CTFG framework. During training, the autoregressive generator produces $G$ candidate feature sequences per input, evaluated by a tri-objective reward ($R_{\text{cls}}$, $R_{\text{inv}}$, $R_{\text{tmp}}$) and optimized via group-relative advantages $\hat{A}^{(g)}$ without a value function. During inference, a single deterministic forward pass (using predicted means $\mu_{i,j}$ only) generates features, which are flattened into $\tilde{z}_i = \text{vec}(z_i)$ and classified by Logistic Regression to produce the predicted label $\hat{y}_i$.}
\label{fig:framework}
\end{figure*}

Fig.~\ref{fig:framework} depicts the complete CTFG framework, highlighting its design for learning domain-generalizable features. The key innovation lies in the training phase, where an autoregressive decoder generates candidate feature sequences per input, framing feature learning as a sequential decision process amenable to reinforcement learning. A tri-objective reward function captures the essential properties of generalizable representations: class separability, cross-user invariance, and temporal coherence. GRPO leverages these rewards to compute group-relative advantages without training a critic network, updating the policy to favor feature generation patterns that yield superior distributional properties. At inference, stochastic sampling is replaced by deterministic mean prediction, ensuring efficient deployment while preserving representation quality. A lightweight logistic regression classifier then maps the extracted features to activity labels.

\subsection{Problem setting and notation}
\label{sec:problem}

Consider \( K \) source users, each with labeled data \( \mathcal{D}_k = \{(x_i^k, y_i^k)\}_{i=1}^{n_k} \), where \( x_i^k \in \mathbb{R}^{l \times d} \) is a sensor sequence of length \( l \) with \( d \) channels, and \( y_i^k \in \{1, \dots, C\} \) is the activity label. Each user \( k \) induces a distinct joint distribution \( P_k(x, y) \), with \( P_i \neq P_j \) for \( i \neq j \) due to physiological and behavioral heterogeneity. The objective is to learn a feature mapping and classifier using only source data that generalizes to unseen target users drawn from distributions \( \{Q_m\}_{m=1}^M \) satisfying \( Q_m \neq P_k \) for all \( m, k \), while the label space and feature space remain the same.

\subsection{Feature extraction as a Markov decision process}
\label{sec:mdp}

The central insight of CTFG is to recast feature extraction---traditionally a single deterministic forward pass---as a sequential decision-making process. Human activities are compositional across time: a gait cycle comprises distinct phases (heel strike, stance, toe-off, swing) whose temporal ordering carries discriminative information that is largely user-invariant, even though amplitude and waveform morphology vary with limb length, body mass, and muscle composition~\cite{wen2023transformers_timeseries}. Standard extractors compress this hierarchy into a fixed vector in a single pass, conflating user-invariant dynamics with user-specific artifacts. An autoregressive process addresses this by constructing the representation token by token, with each token conditioned on all preceding tokens and the encoded input. Early tokens capture coarse, globally stable patterns (periodicity, dominant frequency); later tokens refine with finer details (phase transitions, inter-joint coordination). Crucially, the RL reward is computed only after the complete sequence, providing a holistic quality signal that evaluates distributional properties---inter-class separability and cross-user alignment---rather than optimizing each token against a per-step surrogate loss.

We formalize this as the following MDP.

\begin{defn}[Feature Generation MDP]
\label{def:mdp}
For each input sample \( x_i \in \mathbb{R}^{l \times d} \):
\begin{itemize}
    \item \textbf{Action} at step \( j \): generation of token \( z_{i,j} \in \mathbb{R}^k \), sampled from a parameterized distribution.
    \item \textbf{Transition}: deterministic concatenation \( s_{i,j+1} = (h_i, z_{i,1:j}) \).
    \item \textbf{State} at step \( j \): \( s_{i,j} = (h_i, z_{i,1:j-1}) \), combining the encoder's hidden representation \( h_i \in \mathbb{R}^{l \times d_{\text{model}}} \) with all previously generated tokens.
    \item \textbf{Reward}: computed once after all \( s \) tokens are generated, evaluating the distributional quality of the complete feature batch.
    \item \textbf{Horizon}: fixed at \( s \) (number of feature tokens).
\end{itemize}
\end{defn}

The state at step \( j \) comprises the encoder representation \( h_i \) (a fixed summary of raw input) and the partial sequence \( z_{i,1:j-1} \) (all prior decisions), enabling the policy to adaptively allocate representational capacity. The continuous action space (\( \mathbb{R}^k \)) reflects the nature of feature representations; each token is sampled from a diagonal Gaussian, providing exploration variance for policy gradient estimation during training and deterministic mean prediction during inference. Transitions are deterministic (state augmented by the new token), so trajectory stochasticity resides entirely in action selection, simplifying gradient computation.

The reward is delayed until the complete sequence is generated. This is deliberate: the properties to optimize---cross-user feature clustering, inter-class separation---are properties of the entire batch, not individual tokens. A per-step reward would require decomposing batch-level properties into token-level contributions, which is ill-defined when a single token may aid discrimination but harm invariance. The delayed reward delegates credit assignment to the RL algorithm.

The policy \( \pi_\theta(z_{i,j} \mid x_i, z_{i,1:j-1}) \), parameterized by \( \theta \), generates the complete representation \( z_i = \{z_{i,1}, \dots, z_{i,s}\} \in \mathbb{R}^{s \times k} \) through \( s \) sequential decisions. The objective is:
\begin{equation}
\theta^* = \arg\max_\theta \; \mathbb{E}_{\pi_\theta}\!\left[\mathcal{R}\!\left(\{z_i\}_{i \in \mathcal{B}}, \{h_i\}_{i \in \mathcal{B}}\right)\right],
\label{eq:objective}
\end{equation}
where \( \mathcal{B} \) is a stratified mini-batch covering all source users and activity classes.

\begin{remark}[Why RL over supervised optimization]
Two properties make RL the appropriate paradigm. First, the reward assesses distributional structure (inter-class separation, cross-user alignment) that emerges from the collective batch, not individual token-label pairs. While supervised analogues exist (center loss~\cite{wen2016discriminative}, MMD~\cite{gretton2012kernel}), they require differentiability with respect to each sampled token, excluding non-differentiable metrics such as rank-based separability. RL requires only evaluability: the policy gradient theorem~\cite{williams1992simple} provides gradient estimates through trajectory log-probabilities. Second, stochastic sampling performs adaptive feature-space augmentation, concentrating exploration in reward-favorable regions while avoiding configurations that collapse inter-class boundaries.
\end{remark}

\subsection{Autoregressive feature generation architecture}
\label{sec:architecture}

The policy \( \pi_\theta \) is realized as a Transformer encoder-decoder~\cite{vaswani2017attention} that maps raw sensor sequences to feature token sequences. The choice of Transformer architecture is motivated by its demonstrated capacity for capturing long-range temporal dependencies through self-attention~\cite{wen2023transformers_timeseries}, which is essential for encoding the multi-scale structure of activity signals. We describe each component and its design rationale below.

\subsubsection{Temporal encoder}

A linear projection \( W_{\text{in}} \in \mathbb{R}^{d \times d_{\text{model}}} \) maps the sensor dimension to the model dimension, followed by additive sinusoidal positional encoding~\cite{foumani2024convtran}:
\begin{equation}
h_i = \text{TransformerEncoder}(x_i W_{\text{in}} + \text{PE}(l)),
\label{eq:encoder}
\end{equation}
where \( \text{PE}(t, q) = \sin(t \cdot \omega_q) \) for even \( q \) and \( \cos(t \cdot \omega_q) \) for odd \( q \), with \( \omega_q = 10000^{-2\lfloor q/2 \rfloor / d_{\text{model}}} \). Sinusoidal encoding is preferred over learned embeddings because its continuous frequency progression captures temporal structure at multiple scales---low frequencies encode coarse window position, high frequencies encode local ordering---matching the multi-scale nature of activity signals and generalizing to unseen sequence lengths.

The encoder's multi-head self-attention computes:
\begin{equation}
\text{Attn}(Q, K, V) = \text{softmax}\!\left(\frac{QK^\top}{\sqrt{d_h}}\right) V,
\label{eq:attention}
\end{equation}
where \( Q, K, V \in \mathbb{R}^{l \times d_h} \) are linear projections and \( d_h = d_{\text{model}} / n_{\text{heads}} \). Multiple heads enable specialization in different temporal relationships---local neighborhoods for stride phases, distant positions for global periodicity. The output \( h_i \in \mathbb{R}^{l \times d_{\text{model}}} \) retains full temporal resolution, providing the decoder with position-specific information for selective cross-attention during generation.

\subsubsection{Autoregressive feature decoder}

The decoder generates tokens incrementally, each conditioned on the encoder output and all preceding tokens:
\begin{equation}
[\mu_{i,j},\; \log \sigma_{i,j}] = \text{TransformerDecoder}(h_i, z_{i,1:j-1}),
\label{eq:decoder_output}
\end{equation}
\begin{equation}
z_{i,j} \sim \mathcal{N}\!\left(\mu_{i,j},\; \text{diag}\!\left(\exp(\log \sigma_{i,j})\right)\right),
\label{eq:sampling}
\end{equation}
where \( \mu_{i,j}, \log \sigma_{i,j} \in \mathbb{R}^k \) parameterize a diagonal Gaussian. Each decoder layer applies three operations: (1)~\textit{masked self-attention} enforces causal structure, restricting step \( j \) to tokens \( z_{i,1:j-1} \); (2)~\textit{cross-attention to \( h_i \)} enables selective retrieval from different temporal regions of the encoded input; (3)~\textit{feed-forward layers} project into Gaussian parameter space.

The causal mask induces an implicit curriculum: early tokens, generated with minimal context, must encode broadly useful features, while later tokens benefit from richer context and specialize. The diagonal Gaussian balances expressiveness with tractability---variance captures uncertainty, naturally decreasing in reward-critical dimensions as training progresses and remaining high in reward-neutral dimensions, inducing implicit feature selection. During inference, only the predicted mean \( \mu_{i,j} \) is used, eliminating sampling noise.

This coarse-to-fine generation emerges from the causal conditioning without explicit enforcement: early tokens capture dominant periodicity, later tokens refine phase transitions and inter-segment coordination.

\subsection{Group-Relative Policy Optimization}
\label{sec:grpo_method}

\subsubsection{Limitations of critic-based advantage estimation}

In standard actor-critic methods, advantages are estimated via Generalized Advantage Estimation (GAE)~\cite{schulman2016high}. Letting \(\delta_{j'} = \hat{r}_{j'} + \gamma V_\phi(s_{i,j'+1}) - V_\phi(s_{i,j'})\) denote the temporal-difference residual at step \(j'\), the advantage at step \(j\) is
\begin{equation}
\hat{A}_{i,j}^{\text{GAE}} = \sum_{j'=j}^{s} (\gamma \lambda)^{j'-j}\, \delta_{j'},
\label{eq:gae}
\end{equation}
where \( V_\phi \) is a learned value function and \(\gamma, \lambda \in [0,1]\) are discount and trace-decay parameters.

In the cross-user setting, this introduces two problems. First, \( V_\phi(s_{i,j}) \) must estimate expected future reward, but this depends on which users populate the current mini-batch---a non-stationary target that induces systematic estimation bias. Second, reward scale varies across leave-one-group-out configurations, causing the effective learning rate to fluctuate and destabilizing convergence.

\subsubsection{GRPO formulation}

GRPO resolves both issues by replacing learned value estimation with empirical within-group normalization. For each sample \( x_i \), the policy generates \( G \) independent complete feature sequences:
\begin{equation}
z_i^{(g)} = \big\{z_{i,1}^{(g)}, \dots, z_{i,s}^{(g)}\big\}, \quad g = 1, \dots, G.
\label{eq:group_sampling}
\end{equation}

Each sequence is generated independently by sampling from the policy's output distribution at each step. The independence of the \( G \) sequences is important: it ensures that the group provides an unbiased sample of the reward distribution under the current policy for the given input, enabling statistically valid advantage estimation.

The group-relative advantage for the \( g \)-th sequence is:
\begin{equation}
\begin{aligned}
\hat{A}^{(g)} &= \frac{R^{(g)} - \bar{R}_G}{\hat{\sigma}_G + \epsilon_s}, \quad
\bar{R}_G = \frac{1}{G}\sum_{g=1}^G R^{(g)}, \\
\hat{\sigma}_G &= \sqrt{\frac{1}{G}\sum_{g=1}^G \left(R^{(g)} - \bar{R}_G\right)^2}
\end{aligned}
\label{eq:grpo_advantage}
\end{equation}

The advantage measures how many standard deviations a sequence's reward lies above or below the group mean, making it independent of absolute reward scale (formalized as affine invariance in Proposition~\ref{prop:variance}). The stabilization constant \( \epsilon_s \) bounds amplification when \( \hat{\sigma}_G \approx 0 \).

The complete GRPO loss combines clipped surrogate objectives with group-relative advantages:
\begin{equation}
\mathcal{L}_{\text{GRPO}}(\theta) = -\frac{1}{Gs}\sum_{g=1}^G\sum_{j=1}^s \Big[\ell_j^{(g)} - \beta_{\text{KL}} D_{\text{KL}}\!\big(\pi_\theta \| \pi_{\text{ref}}\big)\Big],
\label{eq:grpo_loss}
\end{equation}
where \( \ell_j^{(g)} = \min\!\big(\rho_{i,j}^{(g)}\hat{A}^{(g)},\; \text{clip}(\rho_{i,j}^{(g)}, 1{-}\epsilon, 1{+}\epsilon)\hat{A}^{(g)}\big) \), \( \rho_{i,j}^{(g)} = \pi_\theta(z_{i,j}^{(g)} \mid s_{i,j}^{(g)}) / \pi_{\theta_{\text{old}}}(z_{i,j}^{(g)} \mid s_{i,j}^{(g)}) \) is the importance sampling ratio, and \( \pi_{\text{ref}} \) is a frozen reference policy. The clipping constrains the trust region, preventing destabilizing updates when the feature distribution shifts. The KL penalty anchors the policy to the reference, preventing drift into degenerate feature regions where reward signals become uninformative.

\subsubsection{Formal analysis}
\label{sec:grpo_analysis}

We now establish the theoretical properties underpinning GRPO's suitability for cross-user HAR.

\begin{proposition}[Properties of group-relative advantage]
\label{prop:variance}
Let \( R^{(1)}, \dots, R^{(G)} \) be i.i.d.\ reward samples. The group-relative advantage \( \hat{A}^{(g)} \) in Eq.~\eqref{eq:grpo_advantage} satisfies:
(i) \( \mathbb{E}[\hat{A}^{(g)}] = 0 \) (zero-centered);
(ii) \( \text{Var}(\hat{A}^{(g)}) \to 1 \) as \( G \to \infty \);
(iii) Affine invariance: \( \hat{A}^{(g)}(aR + b) = \hat{A}^{(g)}(R) \) for constants \( a > 0, b \).
\end{proposition}

\begin{figure}[!t]
\centering
\includegraphics[width=0.8\columnwidth]{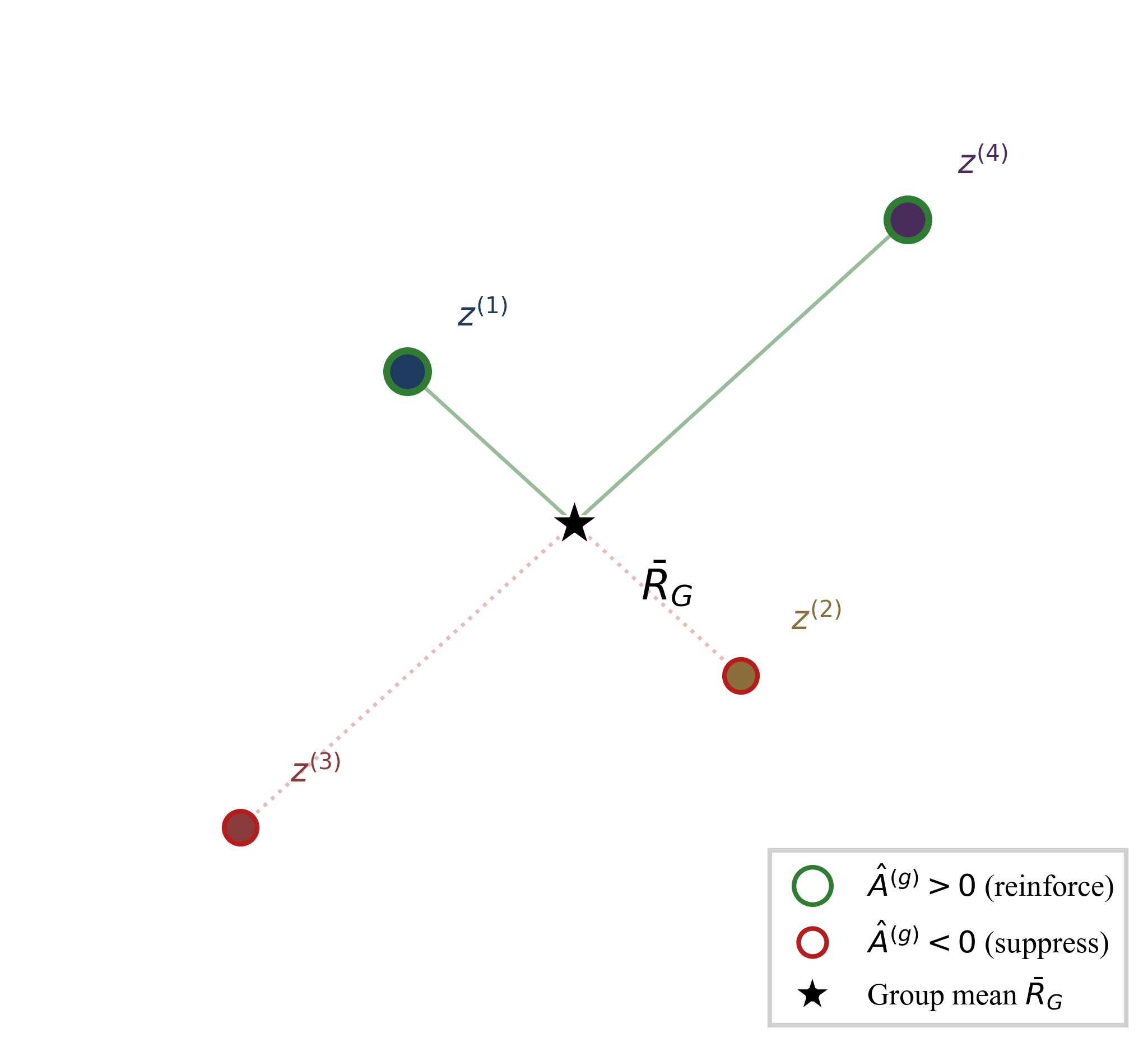}
\caption{Group-relative advantage in the latent space. For each input, $G$ sampled feature sequences are compared against their group mean. Sequences achieving above-average reward receive positive advantages and are reinforced; below-average sequences are suppressed.}
\label{fig:grpo_mechanism}
\end{figure}

Property~(i) ensures the gradient direction reflects relative quality ordering, not absolute reward level. Property~(ii) bounds the advantage magnitude regardless of the underlying reward distribution, stabilizing convergence. Property~(iii) is the most consequential: different source-user combinations produce different reward magnitudes and sensitivities; affine invariance ensures the effective learning rate is constant across leave-one-group-out configurations, eliminating the inter-task variance that plagues critic-based methods. Fig.~\ref{fig:grpo_mechanism} visualizes this mechanism.

\subsection{Tri-objective reward mechanism}
\label{sec:rewards}

The reward comprises three components evaluated over a mini-batch of generated features \( Z = \{z_i\}_{i \in \mathcal{B}} \) and encoder representations \( H = \{h_i\}_{i \in \mathcal{B}} \), each targeting a necessary property of generalizable features.

\subsubsection{Class discrimination reward}
\begin{equation}
R_{\text{cls}}(Z) = \frac{1}{C(C-1)} \sum_{\substack{c, c'=1 \\ c \neq c'}}^C \left\| \bar{\mu}_c - \bar{\mu}_{c'} \right\|_F^2,
\label{eq:rcls}
\end{equation}
where \( \bar{\mu}_c \in \mathbb{R}^{s \times k} \) is the centroid of feature sequences for class \( c \). Centroids are preferred over pairwise sample distances for robustness to stochastic generation outliers and \( O(|\mathcal{B}|) \) vs.\ \( O(|\mathcal{B}|^2) \) complexity---important since the reward is computed \( G \) times per input. The Frobenius norm operates on the full \( s \times k \) matrix, preserving the sequential structure in the discrimination objective.

\subsubsection{Cross-user invariance reward}
\begin{equation}
R_{\text{inv}}(Z) = -\sum_{c=1}^C \Big(\mathcal{V}_c + \mathcal{D}_c\Big),
\label{eq:rinv}
\end{equation}
where \(\mathcal{V}_c\) measures intra-user scatter and \(\mathcal{D}_c\) measures inter-user centroid distance within class \( c \). The two terms address complementary failure modes: tight per-user clusters that are far apart (\(\mathcal{D}_c\) high) indicate user-dependent encoding; loose clusters (\(\mathcal{V}_c\) high) indicate noisy features. Together they enforce a feature space where same-activity features form a single compact cluster irrespective of user identity.

\subsubsection{Temporal fidelity reward}

Without a content-preservation constraint, the policy could achieve high \( R_{\text{cls}} \) and \( R_{\text{inv}} \) by mapping all within-class inputs to a single point, discarding the temporal information needed to disambiguate challenging activity pairs. The temporal fidelity reward prevents this collapse:
\begin{equation}
R_{\text{tmp}}(Z, H) = -\frac{1}{|\mathcal{B}|}\sum_{i \in \mathcal{B}} \left\| W_{\text{proj}} \cdot \overline{z}_i - \overline{h}_i \right\|_2^2,
\label{eq:rtmp}
\end{equation}
where \( \overline{z}_i = \frac{1}{s}\sum_{j=1}^s z_{i,j} \), \( \overline{h}_i = \frac{1}{l}\sum_{t=1}^l h_{i,t} \), and \( W_{\text{proj}} \in \mathbb{R}^{k \times d_{\text{model}}} \) is a learnable projection. The projection is learnable because the feature and encoder spaces have different dimensionalities and semantics. Summary-level matching (via temporal averaging) is less restrictive than point-wise correspondence but sufficient to prevent information collapse.

\subsubsection{Combined objective}

The three rewards are combined as a weighted sum:
\begin{equation}
\mathcal{R}(Z, H) = w_{\text{cls}}\, R_{\text{cls}}(Z) + w_{\text{inv}}\, R_{\text{inv}}(Z) + w_{\text{tmp}}\, R_{\text{tmp}}(Z, H).
\label{eq:total_reward}
\end{equation}

The weights \( w_{\text{cls}}, w_{\text{inv}}, w_{\text{tmp}} \) control the relative importance of each objective. The default configuration (\( w_{\text{cls}} = 3.0, w_{\text{inv}} = 2.0, w_{\text{tmp}} = 1.0 \)) prioritizes class discrimination, reflecting the primary task requirement that activities must be distinguishable. The invariance weight is lower because overly aggressive alignment can erase class-discriminative features that happen to correlate with user identity. The temporal fidelity weight is the smallest, consistent with its role as a regularizer that prevents collapse without dominating the optimization landscape. This weighting hierarchy follows prior research on balancing discrimination and alignment objectives in cross-user HAR~\cite{ye2024deep, ye2025adversarial}.

\subsection{Training algorithm and downstream classifier}
\label{sec:training}

Algorithm~\ref{alg:tprldg_grpo} summarizes the training procedure. Stratified mini-batch sampling (Line~3) ensures representation from all source users and activity classes, which is essential for meaningful computation of both \( R_{\text{cls}} \) (requiring all classes) and \( R_{\text{inv}} \) (requiring all users per class). The \( G \) independent sequences per input (Lines~6--10) provide the empirical reward distribution for advantage computation; larger \( G \) improves accuracy (Proposition~\ref{prop:variance}(ii)) at the cost of memory. A Logistic Regression classifier on flattened features \( \tilde{z}_i = \text{vec}(z_i) \in \mathbb{R}^{sk} \) is trained post-hoc (Line~14); its minimal capacity ensures that performance reflects RL-optimized feature quality rather than classifier expressiveness.

\begin{algorithm}[!t]
\caption{CTFG Training}
\label{alg:tprldg_grpo}
\begin{algorithmic}[1]
\REQUIRE Source data $\{(x_i, y_i, u_i)\}_{i=1}^{N_s}$; weights $w_{\text{cls}}, w_{\text{inv}}, w_{\text{tmp}}$; clip $\epsilon$; group size $G$; KL coeff.\ $\beta_{\text{KL}}$; tokens $s$; epochs $E$
\ENSURE Trained policy $\pi_\theta$
\STATE Init policy $\pi_\theta$ (encoder-decoder), reference $\pi_{\text{ref}} \leftarrow \pi_\theta$, projection $W_{\text{proj}}$
\FOR{epoch $= 1$ to $E$}
    \STATE Sample stratified mini-batch $\mathcal{B}$ covering all users and classes
    \FOR{each $x_i \in \mathcal{B}$}
        \STATE $h_i \leftarrow \text{Encoder}(x_i W_{\text{in}} + \text{PE}(l))$
        \FOR{$g = 1$ to $G$}
            \FOR{$j = 1$ to $s$}
                \STATE $[\mu_{i,j}^{(g)}, \log\sigma_{i,j}^{(g)}] \leftarrow \text{Decoder}(h_i, z_{i,1:j-1}^{(g)})$
                \STATE $z_{i,j}^{(g)} \sim \mathcal{N}(\mu_{i,j}^{(g)}, \text{diag}(\exp(\log\sigma_{i,j}^{(g)})))$
            \ENDFOR
        \ENDFOR
    \ENDFOR
    \STATE Compute rewards $R^{(g)}$ via Eq.~\eqref{eq:total_reward}
    \STATE Compute advantages $\hat{A}^{(g)}$ via Eq.~\eqref{eq:grpo_advantage}
    \STATE Update $\theta, W_{\text{proj}}$ via $\mathcal{L}_{\text{GRPO}}$ (Eq.~\ref{eq:grpo_loss})
\ENDFOR
\STATE Train Logistic Regression on source features $\{(\text{vec}(z_i), y_i)\}$
\end{algorithmic}
\end{algorithm}

\section{Experiment}
\label{sec:experiments}

\subsection{Benchmarks and protocol}
\label{sec:datasets}

Experiments are conducted on \textbf{DSADS} ~\cite{barshan2014recognizing} (Daily and Sports Activities, 8 subjects, 19 activities, 25\,Hz, 9-axis sensors on torso/arms/legs) and \textbf{PAMAP2} ~\cite{reiss2012introducing} (Physical Activity Monitoring, 6 subjects, 11 activities, 100\,Hz, IMUs on chest/wrist/ankle). Table~\ref{tab_datasets_info} summarizes the configurations.

\begin{table}[!t]
\caption{Benchmark dataset configurations.}
\label{tab_datasets_info}
\centering
\small
\setlength{\tabcolsep}{3pt}
\begin{tabular}{ll}
\toprule
\multicolumn{2}{l}{\textbf{PAMAP2} --- 6 subjects, 11 activities, 100\,Hz} \\
\midrule
Groups & A=[1,2], B=[5,6], C=[7,8] \\
Sensors & Chest/wrist/ankle; accel.+gyro. \\
Window & 3\,s (300 samples), 50\% overlap \\
\midrule
\multicolumn{2}{l}{\textbf{DSADS} --- 8 subjects, 19 activities, 25\,Hz} \\
\midrule
Groups & A=[1,2], B=[3,4], C=[5,6], D=[7,8] \\
Sensors & Torso/arms/legs; accel.+gyro. \\
Window & 3\,s (75 samples), 50\% overlap \\
\bottomrule
\end{tabular}
\end{table}

Per-user z-score normalization removes amplitude biases from physiological factors. Leave-one-group-out cross-validation provides rigorous evaluation of cross-user generalization: 4 cross-user transfers for DSADS, 3 for PAMAP2. Classification accuracy on held-out target users is the primary metric. Architecture: 1-layer Transformer, \( d_{\text{model}} = 64 \), 4 heads, lr $= 10^{-4}$ (Adam).

\subsection{Baselines and comparative rationale}
\label{sec:baselines}

The baseline methods are selected to systematically validate distinct aspects of the CTFG framework. Together, they span the spectrum from minimal domain-aware training to sophisticated temporal adaptation to adversarial co-learning, enabling controlled attribution of performance gains to specific design choices.

\textbf{ERM}~\cite{zhang2018mixup} (Empirical Risk Minimization) trains a standard model by minimizing the average empirical loss over all source data without any domain generalization mechanism. It serves as the \emph{lower-bound reference}, quantifying the performance achievable without explicit handling of cross-user distribution shift. The gap between CTFG and ERM isolates the total benefit of the proposed sequential feature generation paradigm.

\textbf{RSC}~\cite{huang2020self} (Representation Self-Challenging) improves generalization by iteratively discarding the dominant features that the model relies upon during training, forcing the network to discover more diverse and robust feature patterns. RSC validates whether \emph{feature diversity alone}---achieved through a self-challenging mechanism that operates on fixed, non-sequential representations---suffices for cross-user generalization, or whether the sequential structure of our autoregressive generation is essential.

\textbf{ANDMask}~\cite{parascandolo2021learning} learns representations by retaining only gradient components that are consistent across all training domains, masking out gradients that point in conflicting directions across users. This invariance-by-gradient-agreement strategy validates whether \emph{implicit invariance constraints} at the gradient level can match the explicit distributional alignment enforced by our cross-user invariance reward. ANDMask represents a fundamentally different approach to invariance---filtering gradients rather than shaping the feature distribution---and its comparison reveals whether distributional reward signals provide stronger invariance guarantees.

\textbf{AdaRNN}~\cite{du2021adarnn} addresses temporal distribution shift by adaptively learning segment-level recurrent weights, re-weighting hidden states across temporal periods to reduce distribution discrepancy. It serves as the \emph{temporal modeling baseline}, validating whether our Transformer-based autoregressive architecture with global self-attention outperforms recurrent temporal adaptation. The comparison is particularly informative because AdaRNN explicitly targets temporal distribution shift---the same phenomenon our framework addresses---but through a fundamentally different mechanism (recurrent adaptation vs.\ autoregressive generation with RL-driven optimization).

\textbf{ACON}~\cite{liuboosting} (Adversarial Co-learning Network) combines adversarial domain alignment with co-learning strategies to bridge distributional discrepancies across domains for cross-domain HAR. It represents the \emph{adversarial alignment paradigm}---the dominant approach in prior cross-user HAR work---and its comparison tests whether our reward-guided generation framework provides more stable and effective alignment than adversarial min-max optimization, which is known to suffer from training instability and mode collapse.

\textbf{PPO-variant} replaces GRPO with PPO+GAE~\cite{schulman2017proximal} within our framework, keeping all other components identical (same encoder-decoder architecture, same tri-objective reward, same training protocol). This serves as the \emph{controlled ablation} for the optimization algorithm, directly isolating the contribution of critic-free advantage estimation versus critic-based estimation. The PPO-variant is the most important baseline because it attributes any performance difference specifically to the GRPO mechanism---the affine-invariant advantage computation and the elimination of the value function---rather than to the autoregressive architecture or the reward design.

\subsection{Cross-user classification results}

\begin{table}[!t]
\caption{Cross-user accuracy (\%) on DSADS.}
\label{tab:dsads}
\centering
\small
\setlength{\tabcolsep}{1.0pt}
\begin{tabular}{lcccccc}
\toprule
\textbf{Method} & \textbf{ABC$\to$D} & \textbf{ACD$\to$B} & \textbf{ABD$\to$C} & \textbf{BCD$\to$A} & \textbf{AVG} & \textbf{STD} \\
\midrule
ERM & 76.56 & 77.25 & 81.03 & 75.39 & 77.56 & 2.11 \\
RSC & 78.92 & 83.94 & 82.47 & 80.42 & 81.44 & 1.92 \\
ANDMask & 82.36 & 78.75 & 84.87 & 82.22 & 82.05 & 2.18 \\
AdaRNN & 79.53 & 77.84 & 83.23 & 76.92 & 79.38 & 2.41 \\
ACON & 85.43 & \underline{88.76} & 89.17 & 87.25 & 87.65 & \underline{1.47} \\
PPO-var. & \underline{87.77} & 85.48 & \underline{91.96} & \underline{87.93} & \underline{88.29} & 2.33 \\
\textbf{Ours} & \textbf{87.22} & \textbf{85.78} & \textbf{93.21} & \textbf{87.91} & \textbf{88.53} & \textbf{2.87} \\
\bottomrule
\end{tabular}
\end{table}

\begin{table}[!t]
\caption{Cross-user accuracy (\%) on PAMAP2.}
\label{tab:pamap2}
\centering
\small
\setlength{\tabcolsep}{3.0pt}
\begin{tabular}{lccccc}
\toprule
\textbf{Method} & \textbf{AB$\to$C} & \textbf{AC$\to$B} & \textbf{BC$\to$A} & \textbf{AVG} & \textbf{STD} \\
\midrule
ERM & 61.42 & 66.38 & 64.75 & 64.18 & \underline{2.06} \\
RSC & 64.27 & 71.46 & 65.13 & 66.95 & 3.21 \\
ANDMask & 58.75 & 67.13 & 68.72 & 64.87 & 4.37 \\
AdaRNN & 62.34 & 74.47 & 65.95 & 67.59 & 5.09 \\
ACON & 67.31 & 75.97 & 72.58 & 71.95 & 3.56 \\
PPO-var. & \underline{69.01} & \underline{79.29} & \underline{74.14} & \underline{74.15} & 4.20 \\
\textbf{Ours} & \textbf{72.10} & \textbf{80.15} & \textbf{73.42} & \textbf{75.22} & \textbf{3.52} \\
\bottomrule
\end{tabular}
\end{table}

Tables~\ref{tab:dsads} and \ref{tab:pamap2} report the cross-user classification results. We organize the analysis around three progressively deeper questions: (1) Does RL-guided sequential feature generation improve over conventional methods? (2) Which specific design components drive the improvement? (3) How do the structural properties of GRPO manifest in cross-task consistency?

\subsubsection{Overall performance and the sequential generation advantage}

CTFG achieves 88.53\% on DSADS and 75.22\% on PAMAP2, gains of 10.97\% and 11.04\% over ERM. These gains quantify the benefit of sequential feature construction over single-pass extraction. ERM optimizes all feature dimensions simultaneously, conflating user-invariant temporal dynamics with user-specific amplitude artifacts. The autoregressive decoder's causal structure creates implicit prioritization: early tokens encode broadly useful features; later tokens refine with finer details conditioned on established coarse structure. This naturally separates stable temporal motifs from user-specific variations, explaining the consistent improvement.

\subsubsection{Disentangling contributions: diversity, invariance, and temporal modeling}

RSC improves over ERM by 3.88\% (DSADS) and 2.77\% (PAMAP2), confirming that feature diversity aids generalization but is insufficient alone. RSC diversifies \emph{which} features are extracted but cannot control \emph{how} they compose across the temporal dimension; CTFG's stochastic group sampling explores different temporal composition strategies, where each \( z_i^{(g)} \) represents a distinct organizational hypothesis.

ANDMask achieves 82.05\% on DSADS but shows high variance on PAMAP2 (STD 4.37). This reveals that gradient-level invariance---retaining only cross-domain gradient agreement---works well when source domains densely cover the user distribution (DSADS: 8 subjects, 4 groups) but fails when coverage is sparse (PAMAP2: 6 subjects, 3 groups). The dramatic AB$\to$C drop to 58.75\% confirms that gradient overlap between groups A and C is insufficient when group B is excluded. CTFG's distributional invariance reward (Eq.~\ref{eq:rinv}) directly penalizes cross-user divergence, avoiding this fragility.

AdaRNN achieves only 67.59\% on PAMAP2 with the highest variance (STD 5.09), providing the most instructive temporal modeling comparison. Both AdaRNN and CTFG address temporal distribution shift, but AdaRNN's recurrent segment-level reweighting is limited by vanishing gradients over PAMAP2's 300-sample windows~\cite{wen2023transformers_timeseries} and by passive feature modulation---it can suppress existing features but cannot compose new ones from temporal relationships. CTFG's Transformer captures long-range dependencies via self-attention, and the decoder actively constructs features by attending to arbitrary temporal positions. The 7.63\% gap reflects the compounding benefit of global attention and active feature construction.

\subsubsection{Adversarial alignment vs.\ reward-guided generation}

ACON achieves competitive performance (87.65\% DSADS, 71.95\% PAMAP2) but relies on adversarial min-max optimization where the discriminator provides per-sample gradients without assessing global distributional structure. This local signal is vulnerable to mode collapse---mapping all features to one point satisfies the adversarial objective but destroys discriminability. CTFG's tri-objective reward evaluates batch-level structure directly: mode collapse minimizes \( R_{\text{cls}} \), user-specific encoding minimizes \( R_{\text{inv}} \), and information loss minimizes \( R_{\text{tmp}} \), forcing the policy toward genuinely generalizable configurations.

The per-transfer pattern is revealing: ACON achieves its best on ACD$\to$B (88.76\%) but drops to 85.43\% on ABC$\to$D, suggesting convergence to a distributional compromise. CTFG's wider range (85.78\%--93.21\%) but higher peak at ABD$\to$C (93.21\%, exceeding ACON's best by 4.04\%) reflects group-relative exploration's ability to discover transfer-specific optima.

\subsubsection{Isolating the GRPO contribution: controlled ablation with PPO}

The PPO-variant shares the identical architecture and reward, differing only in advantage estimation. On DSADS, the 0.24\% average gap (88.29\% vs.\ 88.53\%) masks a per-transfer cross-over: CTFG leads on ABD$\to$C by 1.25\% but trails on ABC$\to$D by 0.55\%, indicating that both optimizers find good solutions on DSADS's well-conditioned landscape.

PAMAP2 sharpens the distinction. The STD drops from 4.20 to 3.52 (16.2\% reduction), directly reflecting affine invariance (Proposition~\ref{prop:variance}(iii)): when the held-out group changes, the reward distribution shifts, and PPO's critic becomes miscalibrated, producing inconsistent updates. GRPO's normalization absorbs these scale changes automatically. The largest single-transfer gain is on AB$\to$C (+3.09\%), PAMAP2's most challenging transfer where groups A and B may share more physiological similarity with each other than with group C (users 7, 8), maximizing the critic's distribution-dependent bias.

\subsection{Convergence and training stability}
\label{sec:convergence}

\begin{figure}[!t]
\centering
\includegraphics[width=\columnwidth]{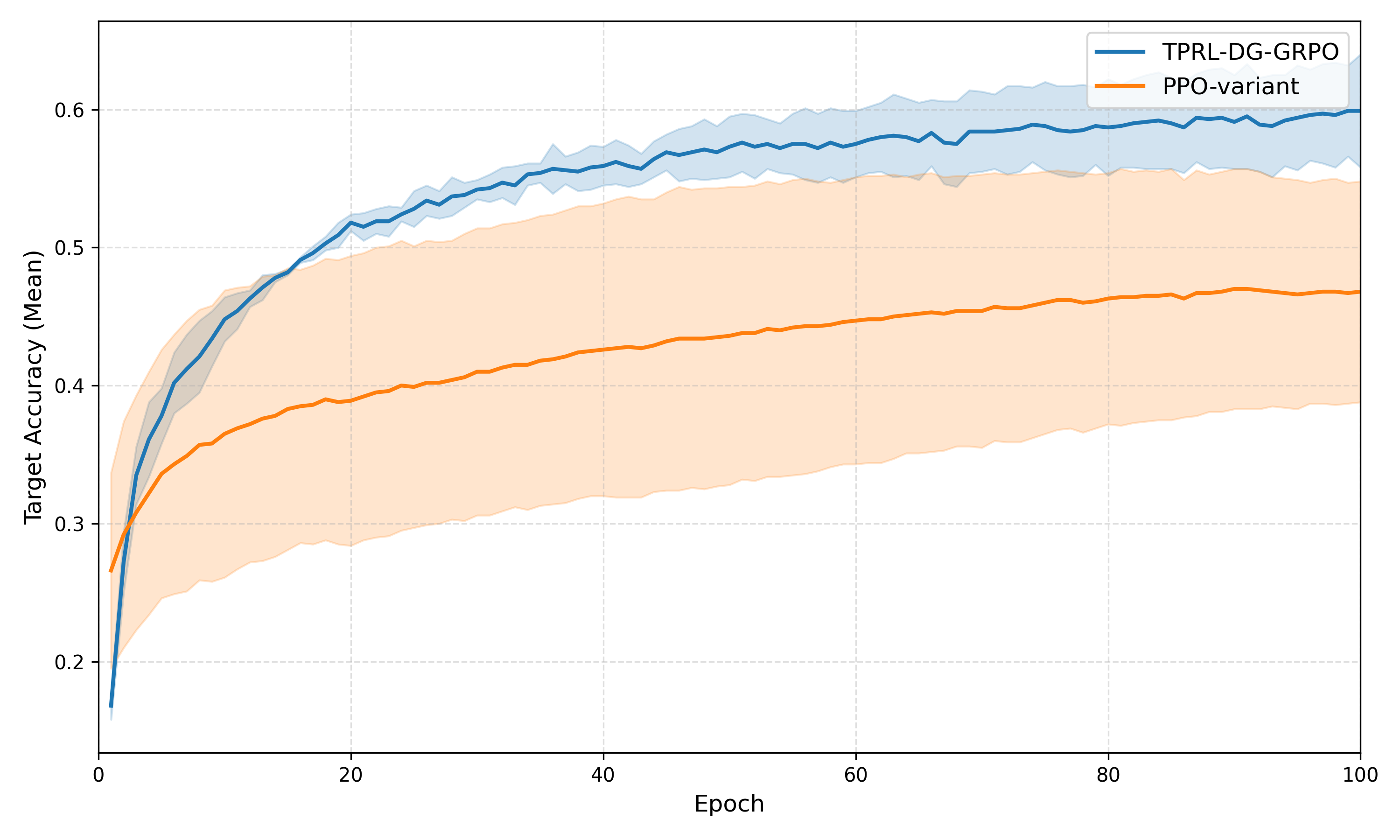}
\caption{Convergence comparison on PAMAP2 (epochs 0--100). Shaded regions indicate $\pm$1 standard deviation across leave-one-group-out configurations.}
\label{fig:conv_pamap2}
\end{figure}

\begin{figure}[!t]
\centering
\includegraphics[width=\columnwidth]{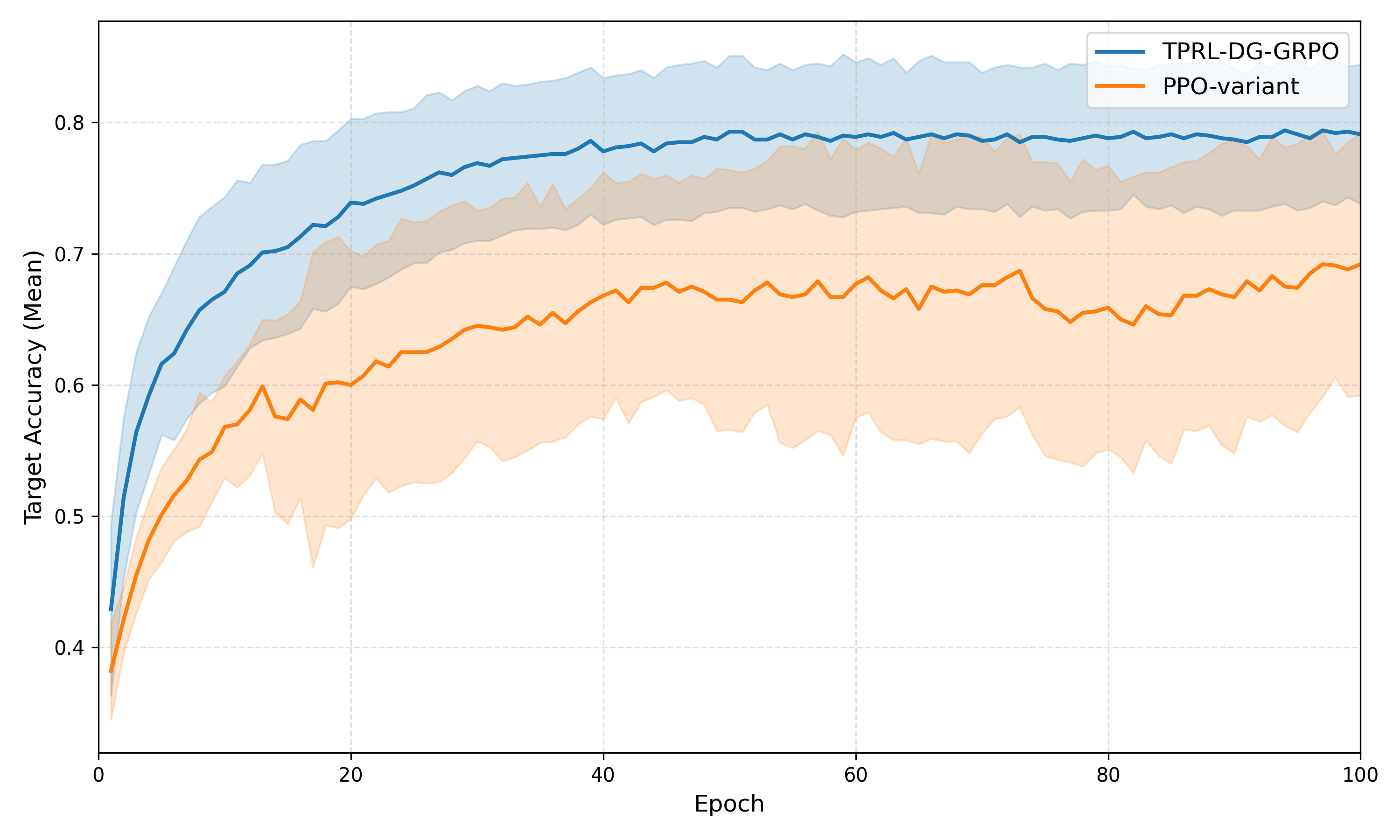}
\caption{Convergence comparison on DSADS (epochs 0--100). GRPO converges faster with monotonically decreasing variance, while PPO exhibits mid-training instability.}
\label{fig:conv_dsads}
\end{figure}

Figures~\ref{fig:conv_pamap2} and \ref{fig:conv_dsads} reveal three phenomena connecting directly to GRPO's design properties.

\textbf{Convergence speed.} On PAMAP2, GRPO reaches 51.8\% at epoch~20---a level PPO does not attain even at epoch~100 (46.8\%). This gap arises because GRPO directly exploits reward variance within each group from epoch~1, while PPO must first learn an accurate value function before producing useful advantages. This ``bootstrap delay'' is particularly costly on PAMAP2 where limited user diversity provides fewer reward patterns for value function training.

\textbf{Inter-task variance.} At epoch~20, GRPO's inter-task STD is 0.016 vs.\ PPO's 0.105 (84.8\% reduction). PPO's value function, trained on one configuration's reward distribution, produces scale-dependent advantages whose magnitude fluctuates across configurations, causing inconsistent effective learning rates. GRPO's normalization (Eq.~\ref{eq:grpo_advantage}) renders advantages scale-invariant, maintaining constant effective learning rate regardless of configuration.

\textbf{Mid-training instability.} On DSADS, PPO's STD increases from 0.168 to 0.212 during epochs 40--60, while GRPO narrows monotonically. This reflects a negative feedback loop: policy improvement shifts the reward distribution, miscalibrating the critic, producing biased updates that temporarily degrade performance. GRPO is immune because advantages are computed fresh from current-group rewards at each iteration, automatically adapting to the evolving reward distribution.

\subsection{Action-space dimensionality analysis}
\label{sec:sensitivity}

\begin{figure}[!t]
\centering
\includegraphics[width=0.6\columnwidth]{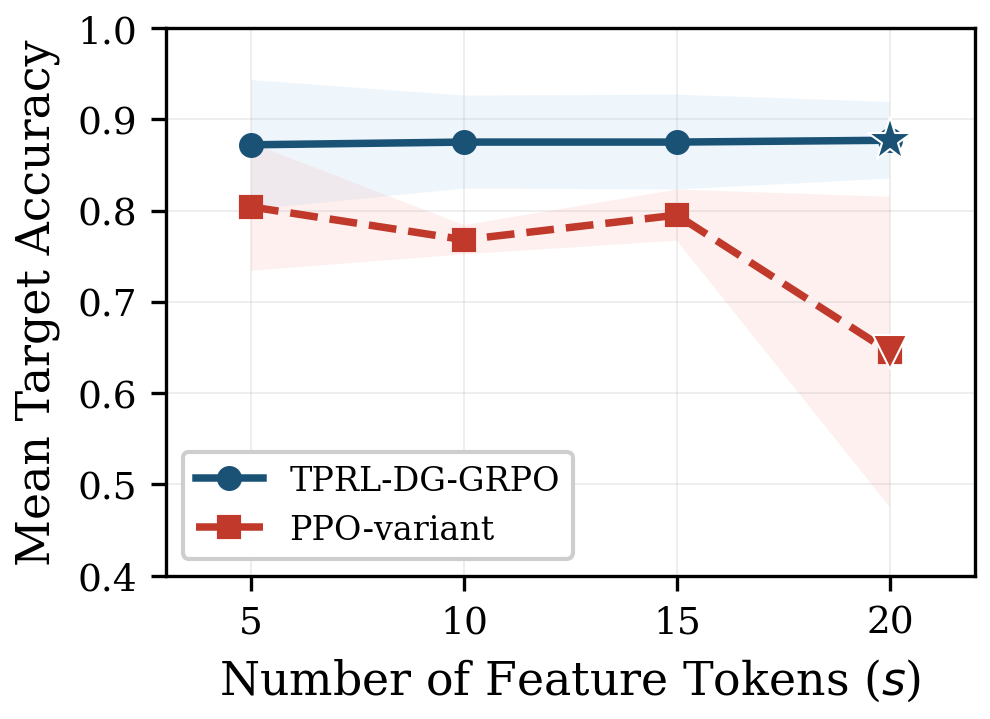}
\caption{Token count sensitivity on DSADS. GRPO maintains stable accuracy with decreasing variance, while PPO collapses at high token counts.}
\label{fig:tok_dsads}
\end{figure}

\begin{figure}[!t]
\centering
\includegraphics[width=0.6\columnwidth]{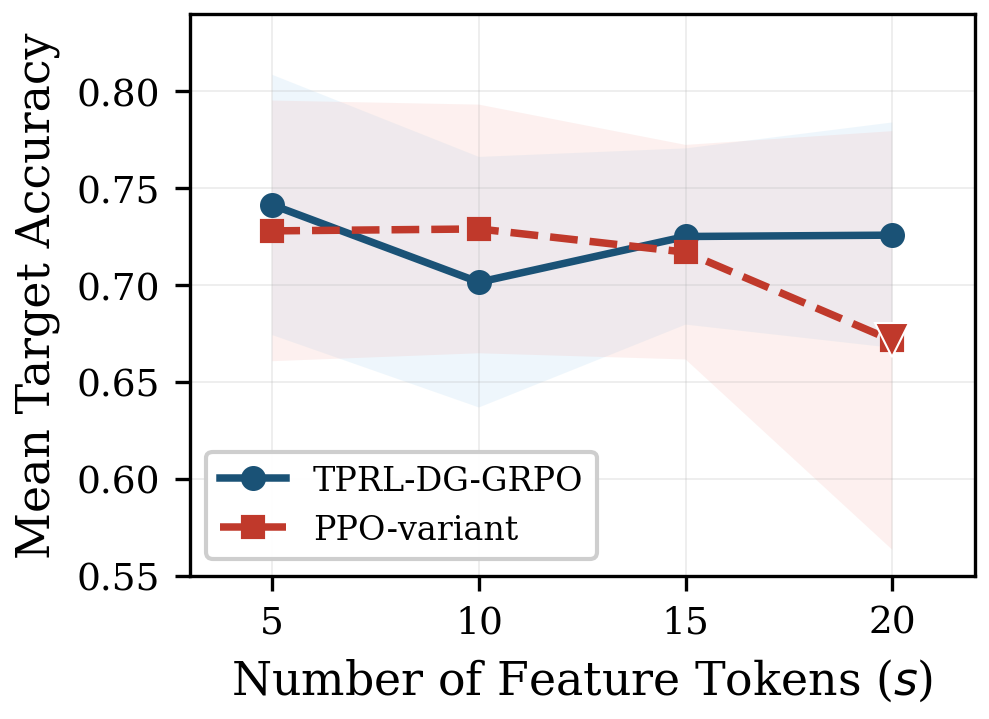}
\caption{Token count sensitivity on PAMAP2. Both methods degrade at $s=20$ on this smaller dataset due to reward landscape flattening.}
\label{fig:tok_pamap2}
\end{figure}

The token count \( s \) controls the generation horizon---the number of sequential decisions the decoder must make. Increasing \( s \) simultaneously increases the representational capacity and the credit assignment difficulty, directly probing the scalability of each optimization paradigm (Figures~\ref{fig:tok_dsads}--\ref{fig:tok_pamap2}).

\textbf{GRPO exhibits remarkable stability on DSADS.} Accuracy remains within a narrow 0.6\% band (87.15\%--87.74\%) across all token counts, while inter-task STD decreases monotonically from 7.05\% at \( s=5 \) to 4.21\% at \( s=20 \). This stability reveals that the autoregressive decoder effectively utilizes additional tokens---each token refines the representation via cross-attention to different temporal regions---without introducing optimization instability. The monotonically decreasing variance is particularly significant: as \( s \) grows, the \( G \) group members sample from a higher-dimensional space, producing greater reward diversity within each group, which improves the statistical quality of GRPO's within-group normalization (Proposition~\ref{prop:variance}(ii)).

\textbf{PPO degrades progressively with increasing horizon.} On DSADS, PPO drops from 80.37\% (\(s=5\)) to 64.46\% (\(s=20\)), a 15.9\% absolute loss that produces a 23.3\% accuracy gap vs.\ GRPO at \(s=20\). The mechanism traces directly to the MDP's sparse reward structure: all intermediate rewards are zero (\(\hat{r}_{j'} = 0\) for \(j' < s\)), so GAE advantages at step \( j \) depend entirely on the value function predicting the final reward from the partial state \( s_{i,j} = (h_i, z_{i,1:j-1}) \). As \( s \) grows, early-step predictions must extrapolate from increasingly incomplete information, and estimation errors at each step compound through the \( s \)-step GAE summation. At \(s=20\), this compounding overwhelms the true advantage signal, producing effectively random gradients. The variance explosion at \(s=20\) (STD surging from 2.76\% at \(s=15\) to 16.97\%) provides direct empirical evidence of this compounding: different leave-one-group-out configurations experience different degrees of value-function miscalibration, producing wildly inconsistent optimization trajectories.

Notably, at intermediate token counts (\(s=10, 15\)), PPO achieves lower inter-task STD than GRPO on DSADS (1.63\% vs.\ 5.05\% at \(s=10\)). This seemingly counterintuitive result has a structural explanation: at moderate horizons, PPO's value function can still produce reasonably accurate estimates, and the critic's smoothing effect reduces the stochastic noise inherent in GRPO's finite-sample group normalization. However, this stability is fragile---it collapses catastrophically once the horizon exceeds the critic's estimation capacity, whereas GRPO's variance reduction is robust and monotonic.

\textbf{PAMAP2 reveals the interaction between horizon and dataset scale.} On PAMAP2, GRPO maintains consistent accuracy across token counts (74.14\% at \(s=5\), 72.58\% at \(s=20\)), with a slight dip at \(s=10\) (70.15\%) that reflects the stochasticity inherent in PAMAP2's smaller sample size (6 subjects, 11 activities). PPO achieves competitive accuracy at low token counts---even slightly outperforming GRPO at \(s=10\) (72.9\% vs.\ 70.15\%)---demonstrating that the autoregressive architecture and tri-objective reward provide substantial benefit regardless of the optimizer when the horizon is short. However, at \(s=20\), PPO drops to 67.14\% with STD exploding to 10.81\% (vs.\ GRPO's 5.82\%), reproducing the same critic-collapse pattern observed on DSADS. The fact that PPO's failure mode manifests identically across both datasets despite their different scales confirms that the root cause is structural (compounding GAE estimation error).

GRPO bypasses this failure mode because its advantage (Eq.~\ref{eq:grpo_advantage}) is computed from the final reward of each complete sequence, assigning a uniform but unbiased signal across all steps. While this is a coarser credit signal than GAE's per-step decomposition, its unbiasedness is the critical property: in the feature generation context, representation quality is a property of the complete token sequence, making sequence-level credit assignment both appropriate and robust.

\subsection{Per-class performance analysis}
\label{sec:perclass}

The per-class analysis tests whether CTFG's performance is predictable from each activity's temporal and kinematic structure, connecting specific patterns to the framework's architecture and reward design.

\begin{table}[!t]
\caption{Per-class performance on PAMAP2.}
\label{tab:perclass_pamap2}
\centering
\small
\setlength{\tabcolsep}{2.5pt}
\begin{tabular}{lrrrrc}
\toprule
\textbf{Activity} & \textbf{Corr.} & \textbf{Tot.} & \textbf{Rec.\%} & \textbf{Prec.\%} & \textbf{F1} \\
\midrule
Lying & 902 & 979 & 92.1 & 89.9 & 0.910 \\
Sitting & 463 & 868 & 53.3 & 68.1 & 0.598 \\
Standing & 578 & 961 & 60.2 & 62.4 & 0.612 \\
Walking & 1067 & 1144 & 93.3 & 73.8 & 0.824 \\
Running & 539 & 632 & 85.3 & 91.2 & 0.881 \\
Cycling & 879 & 937 & 93.8 & 77.9 & 0.851 \\
Nordic walk. & 754 & 1051 & 71.7 & 90.6 & 0.801 \\
Asc.\ stairs & 480 & 599 & 80.1 & 62.9 & 0.705 \\
Desc.\ stairs & 278 & 503 & 55.3 & 59.2 & 0.571 \\
Vacuuming & 675 & 897 & 75.3 & 62.5 & 0.683 \\
Ironing & 770 & 1234 & 62.4 & 87.0 & 0.727 \\
\bottomrule
\end{tabular}
\end{table}

\begin{table}[!t]
\caption{Per-class performance on DSADS.}
\label{tab:perclass_dsads}
\centering
\small
\setlength{\tabcolsep}{2.0pt}
\begin{tabular}{lrrrrc}
\toprule
\textbf{Activity} & \textbf{Corr.} & \textbf{Tot.} & \textbf{Rec.\%} & \textbf{Prec.\%} & \textbf{F1} \\
\midrule
Sitting & 996 & 1206 & 82.6 & 99.1 & 0.901 \\
Standing & 1079 & 1206 & 89.5 & 92.1 & 0.907 \\
Lying on back & 1205 & 1206 & 99.9 & 72.5 & 0.840 \\
Lying on right & 1206 & 1206 & 100.0 & 100.0 & 1.000 \\
Asc.\ stairs & 1150 & 1206 & 95.4 & 99.3 & 0.973 \\
Desc.\ stairs & 1019 & 1206 & 84.5 & 95.1 & 0.895 \\
Stand.\ elev. & 641 & 1206 & 53.2 & 61.2 & 0.569 \\
Walk.\ elev. & 917 & 1206 & 76.0 & 53.4 & 0.627 \\
Walk.\ parking & 951 & 1206 & 78.9 & 95.7 & 0.865 \\
Treadmill (flat) & 995 & 1206 & 82.5 & 85.0 & 0.838 \\
Treadmill (incl.) & 979 & 1206 & 81.2 & 80.8 & 0.810 \\
Running & 1203 & 1206 & 99.8 & 94.7 & 0.971 \\
Stepper & 997 & 1206 & 82.7 & 95.3 & 0.885 \\
Cross trainer & 953 & 1206 & 79.0 & 89.0 & 0.837 \\
Cycling (horiz.) & 1206 & 1206 & 100.0 & 99.5 & 0.998 \\
Cycling (vert.) & 1155 & 1206 & 95.8 & 99.8 & 0.978 \\
Rowing & 1204 & 1206 & 99.8 & 100.0 & 0.999 \\
Jumping & 1151 & 1206 & 95.4 & 97.6 & 0.965 \\
Basketball & 1168 & 1206 & 96.9 & 86.1 & 0.912 \\
\bottomrule
\end{tabular}
\end{table}

\subsubsection{Periodic activities: temporal structure exploitability}

The highest-performing activities share strong repetitive temporal structure: Cycling (horizontal) F1 = 0.998, Rowing 0.999, Running 0.971 on DSADS; Running 0.881, Cycling 0.851 on PAMAP2. These produce periodic signals with biomechanically constrained phase relationships that are largely user-invariant---a tall and short person share the same pedaling phase structure (push-recover-push), differing primarily in amplitude and frequency, both normalized by per-user z-score preprocessing. The autoregressive decoder exploits this systematically: the first token captures the dominant periodic pattern via cross-attention; subsequent tokens refine with progressively higher harmonics (phase asymmetries, transition dynamics), building a multi-resolution representation inherently invariant to user identity.

Lying on right (F1 = 1.000) achieves perfect performance through a different mechanism: its unique gravitational projection across torso/arm/leg sensors shares no overlap with any other activity. The lower F1 for Lying on back (0.840, precision 72.5\%) reveals that its gravitational signature overlaps with some static activities on certain sensor axes, attracting misclassified samples into its feature region.

\subsubsection{Discrimination-invariance tension in kinematically similar pairs}

Walking vs.\ Nordic Walking on PAMAP2 (F1 = 0.824 vs.\ 0.801) exposes the core reward tension. Both share lower-body gait kinematics; only upper-body patterns (arm swing amplitude from pole usage) discriminate them. These discriminative features are precisely the most user-varying (arm swing style is highly personal). The precision-recall asymmetry reveals the resulting decision boundary geometry. Walking achieves high recall (93.3\%) but lower precision (73.8\%); Nordic Walking shows the inverse (71.7\%, 90.6\%). The \( R_{\text{cls}} \) centroid distance maximization pulls Walking toward the generic locomotion region while pushing Nordic Walking to a peripheral region defined by its upper-body signature. Ambiguous instances---Nordic Walking with weak upper-body signal---default to Walking, the larger class.

The DSADS treadmill pair (flat F1 = 0.838, inclined 0.810) exhibits a parallel pattern. Inclination modulates anterior-posterior acceleration and hip flexion in a narrow spectral band that is both subtle and user-dependent. Bidirectional confusion (both recalls $\sim$82\%) confirms the decision boundary bisects a region where user-invariant and class-discriminative features are entangled.

\subsubsection{Static activities and the temporal fidelity floor}

Sitting (F1 = 0.598) and Standing (0.612) on PAMAP2, and Standing in elevator (0.569) on DSADS, represent a principled limitation. These activities produce near-constant signals where the temporal structure the decoder is designed to exploit---periodicity, phase transitions---is absent. Cross-attention queries attend to noise-floor variations that differ across users. Moreover, \( R_{\text{tmp}} \) forces preservation of encoder content that, for static activities, is dominated by the gravitational constant and user-specific noise rather than activity-discriminative dynamics.

The cross-dataset contrast is instructive: DSADS Sitting achieves F1 = 0.901 vs.\ PAMAP2's 0.598. DSADS's torso/arm/leg placement provides a richer multi-axis gravitational signature distinguishing postures even without temporal dynamics (gravity projection across three body segments is posture-specific). PAMAP2's chest/wrist/ankle placement captures the sitting-standing distinction primarily through ankle orientation---a single degree of freedom easily confounded by cross-user postural variation. This confirms that CTFG's static-activity performance is limited by sensor information content.

\subsubsection{Environmental confounding: elevator activities}

Standing in elevator (F1 = 0.569) and Walking in elevator (0.627) are DSADS's lowest-performing activities. The elevator introduces confounding vertical acceleration---independent of the user's activity---that the encoder captures as temporal events and the decoder may interpret as activity dynamics. \( R_{\text{tmp}} \) compounds this by requiring preservation of encoder content that includes confounding signals.

Walking in elevator's low precision (53.4\%) indicates that elevator vibrations produce features resembling low-intensity locomotion, attracting misclassified samples. Standing in elevator's low recall (53.2\%) reflects indistinguishability from regular standing when the elevator is stationary. By contrast, Walking in parking (F1 = 0.865) performs well because the parking environment introduces no confounding temporal dynamics---confirming that the limitation stems from \emph{temporal confounding}, not context-specificity.

\subsubsection{Emergent performance tiers}

A three-tier hierarchy maps onto the framework's architectural properties. \textit{Tier~1} (F1 $\geq$ 0.90): activities with strong periodicity or unique sensor signatures (Cycling, Rowing, Running, Lying on right, Ascending stairs), where all three reward components align constructively. \textit{Tier~2} (F1 0.70--0.89): activities with moderate temporal structure but kinematic overlap (Walking, Nordic Walking, Stepper, Cross trainer), where the discrimination-invariance tension limits performance. \textit{Tier~3} (F1 $<$ 0.70): activities with minimal temporal variation or environmental confounding (Sitting, Standing, elevator activities), representing principled limitations of temporal feature generation.

\section{Conclusion and future work}
\label{sec:conclusion}

This paper has proposed CTFG, a framework that redefines cross-user activity recognition as a reward-guided temporal feature generation process governed by Group-Relative Policy Optimization. The core technical insight is that critic-based RL algorithms introduce a structural bottleneck in the cross-user setting: the value function produces distribution-dependent estimation biases manifesting as inconsistent convergence and high inter-task variance. GRPO eliminates this bottleneck through intra-group reward normalization, yielding self-calibrating, affine-invariant advantage signals. Combined with a tri-objective reward mechanism enforcing class discrimination, cross-user invariance, and temporal fidelity, the framework achieves state-of-the-art accuracy on DSADS (88.53\%) and PAMAP2 (75.22\%), with substantially faster convergence and robust scalability to high-dimensional action spaces where critic-based training collapses.

Future directions include adaptive group sizing strategies for datasets with limited user diversity, class-conditional invariance modulation for kinematically similar activity pairs, computational reduction through importance sampling or amortized advantage estimation, and extension to larger demographically diverse populations and multi-modal sensor configurations.

\bibliographystyle{elsarticle-num}

\end{document}